\documentclass{llncs}

\usepackage[T1]{fontenc}
\usepackage[utf8]{inputenc}
\usepackage{graphicx}
\usepackage{booktabs}
\usepackage{amsmath}

\title{Region-Specific EEG Classification of Hypnotic Susceptibility
       Using a Sliding-Window 2D CNN:\\
       A Subject-Independent Evaluation}

\titlerunning{EEG Classification of Hypnotic Susceptibility via 2D CNN}

\author{Krzysztof Wróbel\inst{1}\orcidID{0000-0002-1452-6623} \and
        Grzegorz M. Wójcik\inst{1}\orcidID{0000-0002-4678-9874}}

\authorrunning{K. Wróbel and G.\,M. Wójcik}

\institute{Department of Neuroinformatics and Biomedical Engineering,\\
           Institute of Computer Science,\\
           Maria Curie-Sk\l{}odowska University, Lublin, Poland\\
           \email{krzysztof.wrobel@umcs.pl, gmwojcik@umcs.edu.pl}}

\begin{document}
\maketitle

\begin{abstract}
We investigate whether a 2D Convolutional Neural Network (CNN) trained on
minimally preprocessed EEG signals can classify hypnotic susceptibility
across distinct brain regions.
Sixty-two male participants (32 Control, 30 Guided Imagery / Hipno group)
underwent a 20-minute EEG recording using a 256-channel EGI cap.
A sliding-window approach evaluates all 10-second windows within the recording
for each of seven predefined brain regions (prefrontal, frontal, temporal,
central, parietal, occipital, and facial), yielding up to 868 independently
trained models.
Initial screening with a segment-level 90/10 random split showed that five
regions exceeded the 0.67 accuracy threshold, with the occipital region
achieving the highest mean accuracy (0.721) across 90.3\% of windows.
Subject-independent evaluation via Leave-One-Subject-Out (LOSO)
cross-validation, however, substantially revised this picture:
occipital performance collapsed to near-chance (mean ACC 0.453, AUC 0.429),
revealing that high random-split accuracy was driven by subject-specific EEG
fingerprints rather than genuine group differences.
In contrast, the frontal and prefrontal regions maintained above-chance LOSO
accuracy (0.646 and 0.624, respectively), demonstrating genuine
generalisability.
These results underscore the necessity of subject-independent validation in
EEG-based classification studies and highlight anterior cortical regions as
more reliable neural markers of hypnotic susceptibility.
\end{abstract}


\section{Introduction}
\label{sec:intro}

Electroencephalography (EEG) is a non-invasive, cost-effective technique
for recording brain bioelectrical activity, increasingly combined with machine
learning to classify mental states~\cite{eeg_sig_proc_and_ml}.
Convolutional Neural Networks (CNNs) have become the dominant deep-learning
architecture in EEG-based classification, achieving strong results in motor
imagery decoding~\cite{xu2019deep}, epileptic seizure detection~\cite{zhou2018epileptic},
emotion recognition~\cite{zhang2020investigation}, and brain-computer
interfaces (BCIs)~\cite{chen2023brain}.

Guided Imagery (GI) is a therapeutic relaxation technique in which structured
mental visualisation modulates physiological and emotional states.
EEG studies of hypnosis-related altered states have documented characteristic
changes in posterior alpha oscillations~\cite{graffin1995eeg,jensen2015mechanisms},
parietal attention networks~\cite{rainville2002hypnosis,landry2017brain},
and transient prefrontal involvement~\cite{egner2005hypnosis}.
Our previous work applied a Generalised Linear Model (GLM) to discriminate
GI and mental-task groups from EEG with 81\% accuracy on a hand-selected
60-second segment~\cite{zemla2023-glm}, and a subsequent CNN comparison
confirmed the advantage of 2D convolutional architectures over recurrent
models for this task~\cite{postepski2025}.

A critical but often overlooked methodological issue is the split strategy
used for evaluation.
When consecutive sub-segments from the same subject are distributed across
training and test sets (segment-level split), the classifier can learn
subject-specific EEG fingerprints~\cite{zhang2022eeg} rather than
group-level differences, inflating apparent accuracy.
Subject-independent Leave-One-Subject-Out (LOSO) cross-validation eliminates
this leakage and provides a more realistic estimate of generalisation.

The present study makes three contributions:
(i)~a systematic sliding-window search over the full 20-minute recording for
    each of seven brain regions, identifying temporally localised windows of
    maximal discriminability;
(ii)~a direct comparison of segment-level and subject-level evaluation
     protocols on the same dataset;
(iii)~identification of the cortical regions whose EEG features genuinely
      generalise across subjects.

\section{Materials and Methods}
\label{sec:methods}

\subsection{Participants and EEG Recording}

Sixty-two healthy male participants aged 19--24 (32 Control, 30 Hipno) were
recruited from Maria Curie-Sk\l{}odowska University (UMCS).
The study was approved by the UMCS Bioethical Commission; all participants
provided written informed consent.
Participants were randomly assigned to the \emph{Hipno group}, who listened to
a 20-minute Guided Imagery recording prepared by a qualified psychologist, or
the \emph{Control group}, who performed a structured mental retrieval task
(recalling Zodiac signs, Polish voivodships, US states, and European capitals)
designed to induce cognitive load~\cite{galy2012relationship}.
Both conditions were delivered in a darkened room with participants lying down
and eyes closed to minimise artefacts.

\begin{figure}[t]
  \centering
  \includegraphics[width=0.80\linewidth]{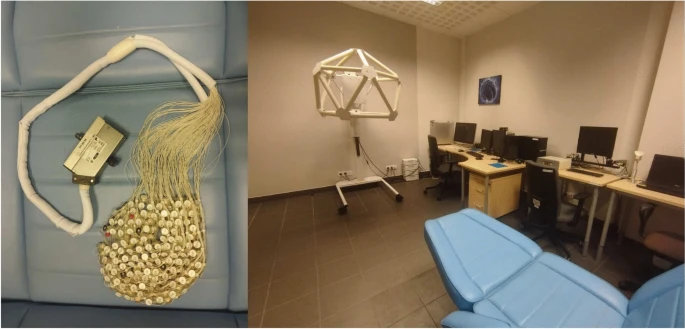}\\[4pt]
  \includegraphics[width=0.52\linewidth]{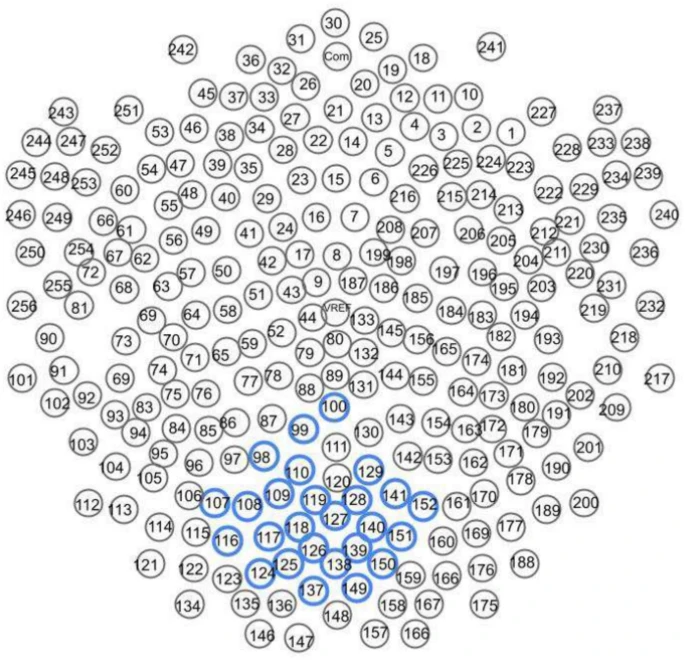}
  \caption{EEG laboratory at UMCS (top) and the 256-channel EGI GSN-130 cap (bottom).}
  \label{fig:lab}
\end{figure}

EEG was recorded at 250~Hz using a 256-channel EGI GSN-130 dense-array cap
and EGI Net Station 4.5.4 software (Fig.~\ref{fig:lab}).

\subsection{Preprocessing}

EEG signals were processed with MNE-Python~\cite{mnePython}.
Bad channels were detected via the RANSAC algorithm~\cite{appelhoff_2023_10047462pyprep}
and reconstructed by interpolation.
Each recording was bandpass filtered to 1--40~Hz, then z-score normalised
per channel over the full recording:
$\hat{x} = (x - \mu)/\sigma$.

The usable recording segment spans 45--1245~s (1200~s total), partitioned
into non-overlapping 10-second windows.
Each window is further divided into 50-sample sub-segments (0.2~s at 250~Hz),
which serve as individual training samples for the CNN.
All 124 windows were evaluated independently for each of seven anatomically
defined brain regions (Table~\ref{tab:regions}), yielding up to
$124 \times 7 = 868$ independently trained models.
Parallelism was implemented with Python's \texttt{asyncio} and
\texttt{ProcessPoolExecutor}.

\begin{table}[t]
  \caption{Brain regions and electrode counts (EGI 256-channel GSN-130 cap).}
  \label{tab:regions}
  \centering
  \begin{tabular}{lc}
    \toprule
    \textbf{Region} & \textbf{Electrodes ($N$)} \\
    \midrule
    Prefrontal & 22 \\
    Frontal    & 20 \\
    Temporal   & 72 \\
    Central    & 25 \\
    Parietal   & 41 \\
    Occipital  & 40 \\
    Facial     & 30 \\
    \bottomrule
  \end{tabular}
\end{table}

\subsection{CNN Architecture}

Each sub-segment is treated as a 2D matrix (time samples $\times$ electrodes),
allowing $3\times3$ convolutional kernels to capture joint spatial-temporal
patterns.
The network consists of two convolutional blocks:
\emph{Block~1} — Conv2D(32 filters, $3\times3$, ReLU), MaxPool2D($2\times2$),
Dropout(0.25);
\emph{Block~2} — Conv2D(64 filters, $3\times3$, ReLU), MaxPool2D($2\times2$),
Dropout(0.25);
followed by Flatten, Dense(64, ReLU), Dropout(0.50), and a Softmax output
layer with 2 units (HIPNO / CONTROL).
The model is compiled with the Adam optimiser and sparse categorical
cross-entropy loss, trained for 20 epochs with batch size 32.

\subsection{Evaluation Protocol}

\textbf{Segment-level screening.}
For each window--region combination, the dataset is split 90\%/10\%
(random seed 42) at the sub-segment level.
A window is labelled \emph{discriminative} when test accuracy $>0.67$.
This fast screening identifies candidate regions and windows but does not
guarantee subject-independent generalisation.

\textbf{Subject-independent evaluation (LOSO).}
For each of the five discriminative regions, LOSO cross-validation was applied
to the single best-performing window of that region.
In each of the 62 folds, the model was trained on 61 subjects and evaluated
on all sub-segments of the held-out subject.
Per-fold accuracy is reported as mean~$\pm$~SD; aggregate F1 and
Area Under the ROC Curve (AUC) are computed from concatenated predictions
of all folds.

\section{Results}
\label{sec:results}

\subsection{Segment-Level Screening}

Table~\ref{tab:random_split} reports the screening results across all seven
regions.
Five regions produced at least one discriminative window.
The occipital region dominated: 90.3\% of windows exceeded the 0.67 accuracy
threshold, with mean accuracy of 0.721 across discriminative windows.
The parietal region ranked second (76.6\%, mean ACC 0.709).
Anterior regions (prefrontal, frontal, central) showed substantially lower
proportions of discriminative windows (10--20\%) but reached comparable peak
accuracies, with the prefrontal achieving the overall highest single-window
accuracy (0.822).
The temporal and facial regions produced no discriminative windows.

\begin{table}[t]
  \caption{Segment-level screening results (90/10 random split). A window is
  discriminative when ACC $>0.67$. Mean ACC is computed over discriminative
  windows only.}
  \label{tab:random_split}
  \centering
  \begin{tabular}{lcccc}
    \toprule
    \textbf{Region} & \textbf{Discriminative} & \textbf{Peak ACC}
                    & \textbf{Mean ACC} & \textbf{Best window (s)} \\
    \midrule
    Occipital  & 112/124 (90.3\%) & 0.803 & 0.721 & 771--781  \\
    Parietal   &  95/124 (76.6\%) & 0.773 & 0.709 & 1121--1131 \\
    Prefrontal &  24/124 (19.4\%) & 0.822 & 0.709 & 1101--1111 \\
    Central    &  20/124 (16.1\%) & 0.803 & 0.700 & 1101--1111 \\
    Frontal    &  13/124 (10.5\%) & 0.783 & 0.711 & 1121--1131 \\
    Temporal   &   0/124  (0.0\%) & ---   & ---   & --- \\
    Facial     &   0/124  (0.0\%) & ---   & ---   & --- \\
    \bottomrule
  \end{tabular}
\end{table}

\subsection{Subject-Independent Evaluation (LOSO)}

Table~\ref{tab:loso} presents LOSO results for the five discriminative
regions.
The most striking finding is the collapse of occipital performance:
mean LOSO accuracy dropped to 0.453 with AUC 0.429 --- below the 0.5 chance
level --- indicating that the high random-split accuracy reflected
subject-specific EEG fingerprints, not group-level differences.
In contrast, the frontal region achieved the highest LOSO mean accuracy
(0.646, AUC 0.692), followed by parietal (0.631, AUC 0.680) and prefrontal
(0.624, AUC 0.681).
These three regions maintained consistently above-chance performance,
confirming genuine cross-subject discriminability.

\begin{table}[t]
  \caption{Leave-One-Subject-Out (LOSO) results for the best window of each
  discriminative region (62 folds). HIPNO group is the positive class.}
  \label{tab:loso}
  \centering
  \begin{tabular}{lcccc}
    \toprule
    \textbf{Region} & \textbf{Mean ACC $\pm$ SD} & \textbf{F1}
                    & \textbf{AUC} & \textbf{Best window (s)} \\
    \midrule
    Frontal    & $0.646 \pm 0.302$ & 0.638 & 0.692 & 1121--1131 \\
    Parietal   & $0.631 \pm 0.228$ & 0.627 & 0.680 & 1121--1131 \\
    Prefrontal & $0.624 \pm 0.353$ & 0.616 & 0.681 & 1101--1111 \\
    Central    & $0.591 \pm 0.243$ & 0.582 & 0.599 & 1101--1111 \\
    Occipital  & $0.453 \pm 0.235$ & 0.434 & 0.429 & 771--781   \\
    \bottomrule
  \end{tabular}
\end{table}

\begin{figure}[t]
  \centering
  \includegraphics[width=\linewidth]{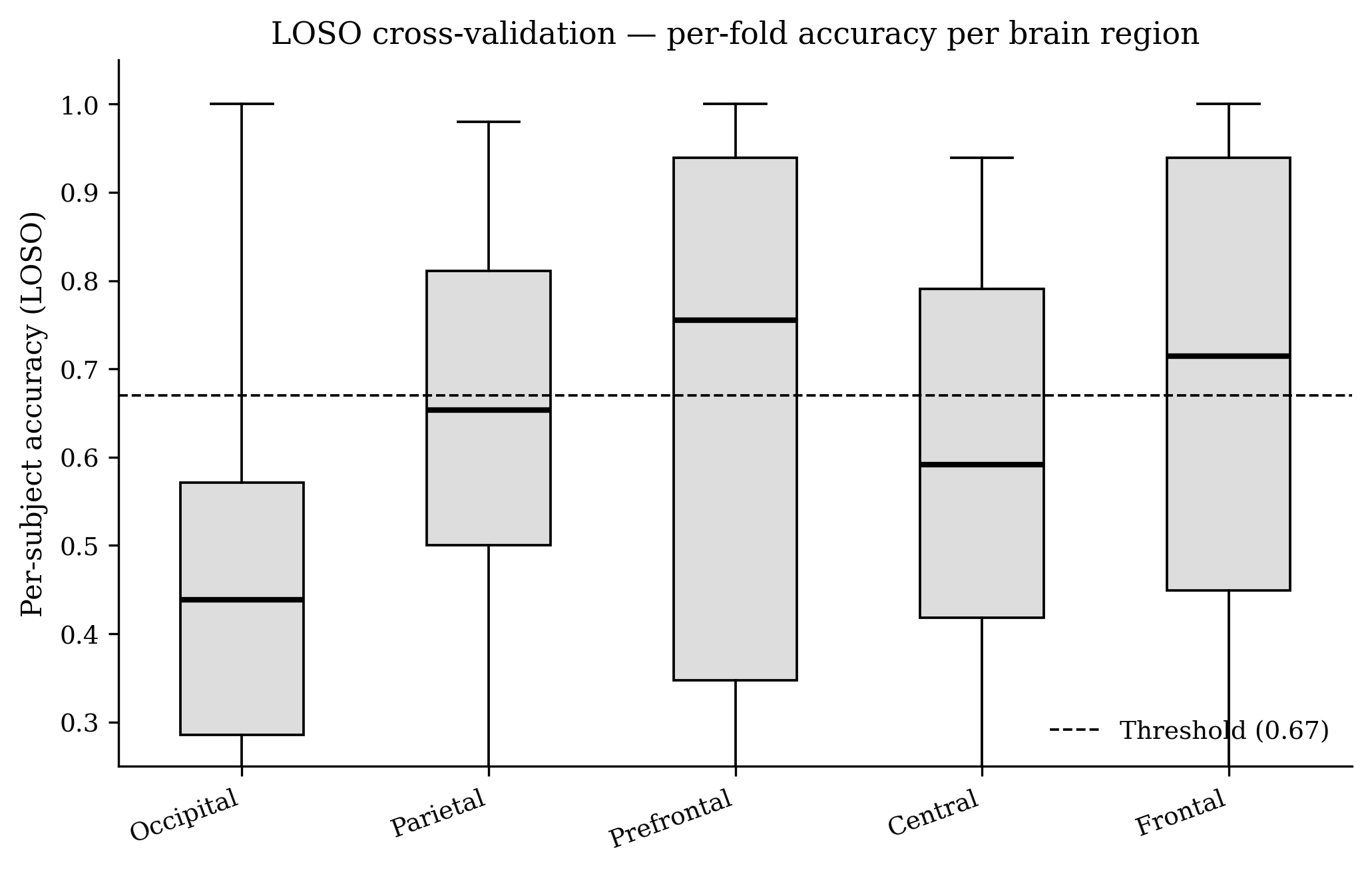}
  \caption{LOSO per-subject accuracy distributions (62 folds) for the five
  discriminative regions. Dashed line marks the 0.5 chance level.}
  \label{fig:loso_box}
\end{figure}

Figure~\ref{fig:loso_box} shows the per-subject accuracy distributions.
The high standard deviations (up to 0.353 for prefrontal) reflect the
considerable inter-subject variability typical of EEG studies with small
cohorts; some subjects are classified reliably while others fall near chance.

\section{Discussion}
\label{sec:discussion}

\textbf{Subject fingerprinting in the occipital region.}
The collapse of occipital accuracy under LOSO (AUC 0.429) is consistent with
known high within-session stability of posterior alpha rhythms, which serve as
effective biometric identifiers~\cite{zhang2022eeg}.
When segments from the same 20-minute session appear in both training and test
sets, a classifier readily learns individual spectral profiles rather than
group membership.
This effect is particularly pronounced in the occipital region, where alpha
power is both highly stable within subjects and strongly modulated by Guided
Imagery~\cite{graffin1995eeg}.
The finding underscores that EEG studies relying solely on segment-level
splits may substantially overestimate classification performance.

\textbf{Anterior and parietal regions as genuine markers.}
The frontal and prefrontal regions, which showed modest proportions of
discriminative windows in the screening phase (10--20\%), turned out to be
the most reliable cross-subject classifiers (AUC 0.692 and 0.681).
This is consistent with the neuroimaging literature on hypnosis: parietal
networks supporting attentional reorientation and multisensory integration
are robustly modulated across individuals~\cite{rainville2002hypnosis,landry2017brain},
whereas prefrontal involvement reflects decoupling of cognitive control during
hypnotic induction~\cite{egner2005hypnosis}.
These features appear less dominated by individual-level variation and thus
carry more generalisable group-level information.

\textbf{Comparison with previous work.}
The LOSO accuracies (0.624--0.646 for the best-generalising regions) are
lower than the 81\% reported by our GLM study~\cite{zemla2023-glm}, which
used a segment-level split on a manually selected 60-second window.
Applying a subject-independent protocol removes the performance inflation due
to segment leakage and provides a more conservative but realistic baseline.
The 2D CNN without hand-crafted features achieves comparable performance to
SVM-based approaches applied to raw EEG~\cite{svmRelaxation}, validating
end-to-end learning as a viable alternative.

\textbf{Limitations.}
The study includes 62 subjects, exclusively male and aged 19--24,
limiting generalisability to other demographics.
No artefact rejection beyond bad-channel interpolation was applied; however,
the absence of discriminative windows in the facial region suggests that
peripheral muscle artefacts did not introduce systematic group differences.

\section{Conclusions}
\label{sec:conclusions}

We demonstrated that a sliding-window 2D CNN applied to minimally preprocessed
256-channel EEG can identify temporally localised windows where hypnotic
susceptibility is discriminable at the segment level in five of seven brain
regions.
Critically, LOSO cross-validation revealed that the dominant occipital
signal reflects subject-specific EEG fingerprints rather than group-level
neural markers.
The frontal and parietal regions, while less prominent in the segment-level
analysis, generalise across subjects and represent more reliable targets for
hypnosis-state classification.
These findings support the use of subject-independent evaluation as a
standard practice in EEG-based classification studies.
Future work should investigate larger and more diverse cohorts and explore
ensemble methods that aggregate predictions across multiple temporal windows
to improve cross-subject robustness.

\section*{Acknowledgements}
The experiments were conducted in the EEG Laboratory of the Department of
Neuroinformatics and Biomedical Engineering at Maria Curie-Sk\l{}odowska
University. The study was approved by the UMCS Bioethical Commission dated 09/07/2021 (approval code: 76/WSP/20).

\bibliographystyle{splncs04}
\bibliography{references}

\end{document}